\documentclass[10pt]{article}

\usepackage{amsmath,graphicx}
\usepackage{amssymb}
\usepackage{amsmath}
\usepackage{cite}
\usepackage{amsthm}
\usepackage{float}
\usepackage{color,subcaption}
\usepackage{bm}
\usepackage{bbm}
\usepackage{algorithm}
\usepackage{algpseudocode}
\usepackage{enumitem}
\usepackage[margin=1in]{geometry}
\usepackage{enumitem}
\usepackage{lipsum}
\usepackage{multirow}
\usepackage{multicol}
\usepackage{booktabs}  
\usepackage{threeparttable}

\def\defn{\,\triangleq\,}

\def\xbf{{\mathbf{x}}}
\def\ybf{{\mathbf{y}}}

\def\ybf{{\mathbf{y}}}

\def\xbf{{\mathbf{x}}}
\def\ybf{{\mathbf{y}}}

\def\Hbf{{\mathbf{H}}}

\def\argmin{\mathop{\mathsf{arg\,min}}}

\theoremstyle{definition}

\graphicspath{{figures/}}

\begin{document}

\title{Image Restoration using Total Variation Regularized Deep Image Prior}


\author{Jiaming~Liu$^1$,~Yu~Sun$^2$, Xiaojian~Xu$^2$ and~Ulugbek~S.~Kamilov$^{1,2}$\\
\small $^1$\emph{Department of Electrical and Systems Engineering,~Washington University in St.~Louis, MO 63130, USA.}\\
\small $^2$\emph{Department of Computer Science and Engineering,~Washington University in St.~Louis, MO 63130, USA.}\\
}


\date{}
\maketitle 


\begin{abstract}
In the past decade, sparsity-driven regularization has led to significant improvements in image reconstruction. Traditional regularizers, such as total variation (TV), rely on analytical models of sparsity. However, increasingly the field is moving towards trainable models, inspired from deep learning. Deep image prior (DIP) is a recent regularization framework that uses a convolutional neural network (CNN) architecture without data-driven training. This paper extends the DIP framework by combining it with the traditional TV regularization. We show that the inclusion of TV leads to considerable performance gains when tested on several traditional restoration tasks such as image denoising and deblurring.
\end{abstract}

\section{Introduction}
\label{Sec:Intro}

Image reconstruction is one of the most widely studied problems in computational imaging. Since the problem is often ill-posed, the process is traditionally regularized by constraining the solutions to be consistent with our prior knowledge about the image. Some traditional imaging priors include nonnegativity, transform-domain sparsity, and self-similarity~\cite{Rudin.etal1992, Figueiredo.Nowak2001, Elad.Aharon2006, Danielyan.etal2012}. Recently, however, the attention in the field has been shifting towards new imaging formulations based on deep learning~\cite{LeCun.etal2015}. 

The most common deep-learning approach is based on an end-to-end training of a convolutional neural network (CNN) for reproducing the desired image from its noisy measurements~\cite{Mousavi.etal2015, Jin.etal2017a, Han.etal2017, Sun.etal2018, Lee.etal2018}. A popular alternative considers training a CNN as an image denoiser and using it within an iterative reconstruction algorithms~\cite{Meinhardt.etal2017, Zhang.etal2017a, Romano.etal2017, Sun.etal2018a}. However, recently, it was also shown that a CNN can by itself regularize image reconstruction \emph{without} data-driven training~\cite{Ulyanov.etal2018}. This \emph{deep image prior (DIP)} framework naturally regularizes reconstruction by optimizing the weights of a CNN for it to synthesize the measurements from a given random input vector. The intuition behind DIP is that natural images can be well represented by CNNs, which is not the case for the random noise and certain other image degradations. DIP was shown to achieve remarkable performance on a number of image reconstruction tasks~\cite{Ulyanov.etal2018, VanVeen.etal2018}.

In this paper, we propose to further improve DIP by combining an \emph{implicit} CNN regularization with an \emph{explicit} TV penalty. The idea of our DIP-TV approach is simple: by including an additional TV term into the objective function, we restrict the solutions synthesized by CNN to those that are piecewise smooth. We experimentally show that our DIP-TV method outperforms the traditional formulations of DIP and TV, and performs on a par with other state-of-the-art image restoration methods such as BM3D~\cite{Dabov.etal2007} and IRCNN~\cite{Zhang.etal2017a}.
\begin{figure*}[t]
\begin{center}
\includegraphics[width=16.7cm]{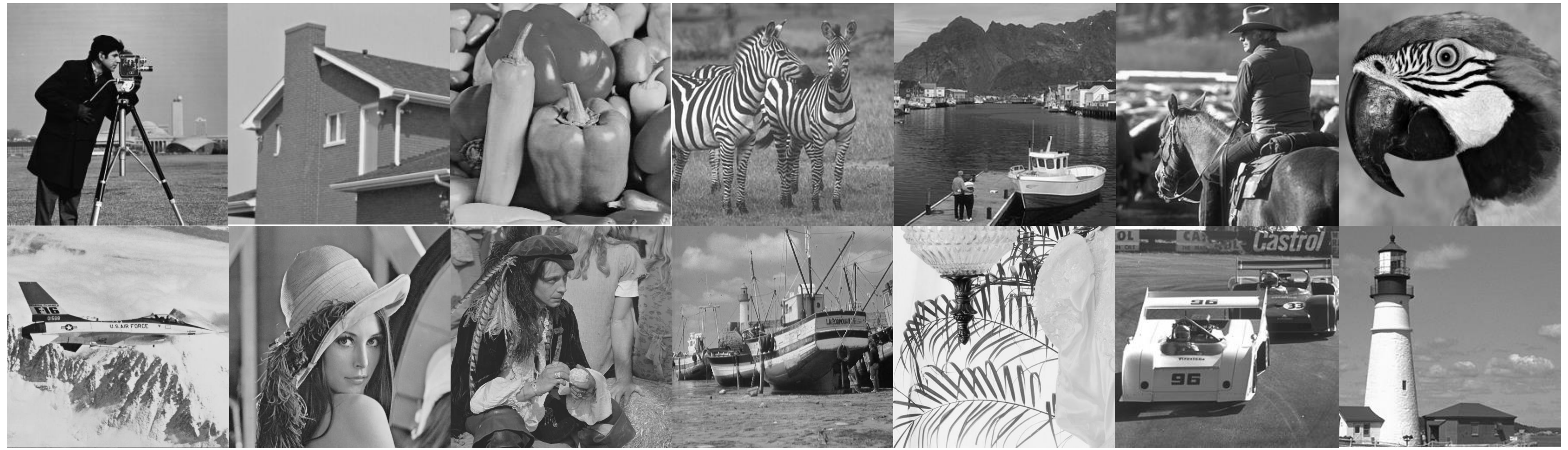}
\end{center}
\caption{The set of 14 grayscale images used in experiments.}
\label{Fig:testimages}
\end{figure*}
\begin{figure}[t]
\begin{center}
\includegraphics[width=16.7cm]{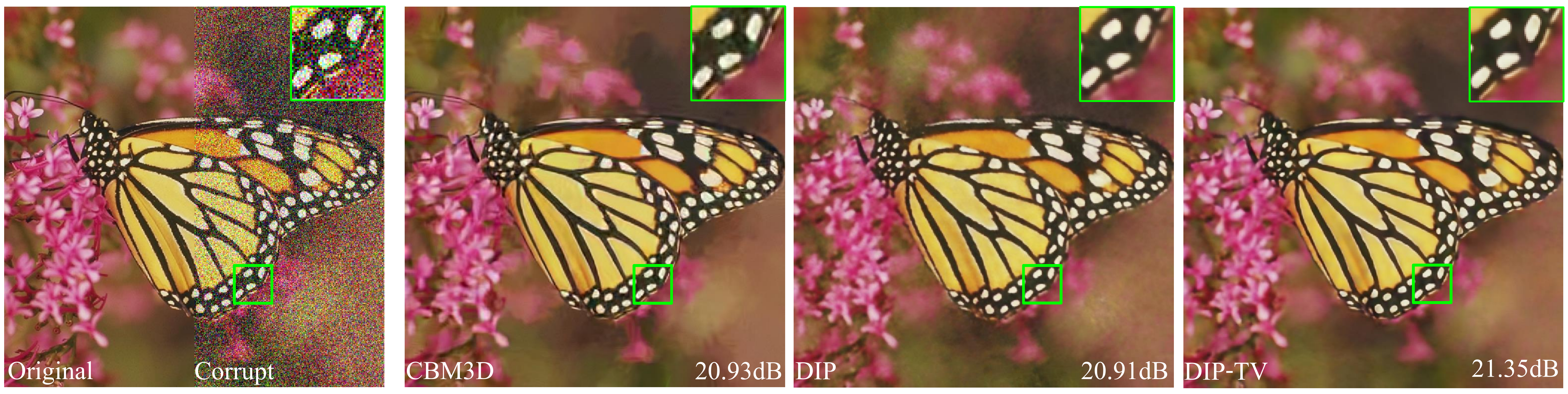}
\end{center}
\caption{Comparison of DIP-TV against several standard algorithms for image denoising. DIP-TV achieves the best SNR performances on image \emph{Monarch} with AWGN of $\sigma$ = 65. The combination of the CNN and TV priors preserve homogeneity of the background as well as the texture, highlighted by rectangles drawn inside the images. }
\label{Fig:color testimage}
\end{figure}
\section{Background}
\label{Sec:Backround}
Consider the restoration as a linear inverse problem
\begin{equation}
\label{Eq:linear inverse problem}
\ybf = \Hbf\xbf + \mathbf{e},
\end{equation}
where the goal is to reconstruct an unknown image $\xbf \in \mathbb{R}^N$ from the measurements $\ybf \in \mathbb{R}^M$. Here, $\Hbf \in \mathbb{R}^{M\times{}N}$ is a degradation
matrix and $\mathbf{e} \in \mathbb{R}^M$ corresponds to the measurement noise, which is assumed to be additive white Gaussian (AWGN) of variance $\sigma^2$. 

As practical inverse problems are often ill-posed, it is common to regularize the task by constraining the solution according some prior knowledge.
In practice, the reconstruction often relies on the regularized least-squares formulation
\begin{equation}
\begin{aligned}
\xbf^\ast = \argmin_{\xbf}\:\left\{\|\ybf-\mathbf{H}\xbf \|_{\ell_2}^2+\lambda\rho (\xbf) \right\}
\end{aligned}
\label{Eq:optimization problem}
\end{equation}
where the data-fidelity term ensures the consistency with measurements, and regularizer $\rho$ constrains the solution to the desired image class. The parameter $\lambda>0$ controls the strength of regularization. 

Total variation (TV) is one of the most widely used image priors that promotes sparsity in image in image gradients~\cite{rudin1992nonlinear}. It has been shown to be effective in a number of  applications ~\cite{Persson.etal2001,Lustig.etal2007,kamilov2016optical}. The $\ell_1$-based anisotropic TV is given by
\begin{equation}
\label{Eq:anisotrotpic TV}
\rho_{TV}(\xbf)\:\:\defn \sum_{\substack{i=1}}^N |[\mathbf{D_1\xbf}]_n| + |[\mathbf{D_2\xbf}]_n|,\quad\quad
\end{equation}
where $\mathbf{D_1}$ and $\mathbf{D_2}$ denote the finite difference operation along the first and second dimension of a two-dimensional (2D) image with appropriate boundary conditions. 

Currently, deep learning achieves the state-of-the-art performance for different image restoration problems~\cite{egmont2002image,xie2012image,Zhang.etal2017}. The core idea is to train a CNN via the following optimization
\begin{equation}
\begin{aligned}
\label{Eq:CNNloss}
&\mathbf{\Theta}^\ast = \argmin_{\mathbf{\Theta}}\:\mathcal{L}(f_{\mathbf{\Theta}}(\ybf),\xbf),\\ 
&\text{such that}\quad\xbf^\ast\: =\:f_{\mathbf{\Theta}^\ast}(\ybf), 
\end{aligned}
\end{equation}
where $\xbf^\ast$ is the restored image, and $f_\mathbf{\Theta}(\cdot)$ represents the CNN parametrized by $\mathbf{\Theta}$. $\mathcal{L}$ denotes the loss function. In practice, ($\ref{Eq:CNNloss}$) can be effectively optimized using the family of stochastic gradient descend (SGD) methods, such as adaptive moment estimation (ADAM)~\cite{kingma2014adam}.

Recently, Ulyanov \emph{et al.}~\cite{Ulyanov.etal2018} proposed to use CNN-based methods in an alternative way. They discovered that the architecture of deep CNN models is well-suited for representing natural images, but not random noise. With a random input vector, CNN can reproduce the clear image without supervised training on a large dataset. In the context of image restoration, the associated optimization for DIP can be formulated as
\begin{equation}
\label{Eq:DIP}
\begin{aligned}
&\mathbf{\Theta}^* = \argmin_{\mathbf{\Theta}}\:\|\ybf-\mathbf{H}f_{\mathbf{\Theta}}(\mathbf{z})\|_{\ell_2}^2,\\
&\text{such that}\quad\xbf^\ast\:=\:f_{\mathbf{\Theta}^*}(\mathbf{z}).
\end{aligned}
\end{equation}
where $\mathbf{z}\in\mathbb{R}^N$ denotes the random input vector. The CNN generator is initialized with random variables $\mathbf{\Theta}$, and these variables are iteratively optimized so that the output of the network is as close to the target measurement as possible.
\section{Proposed Method}
\label{Sec:Method}
\begin{figure}[t]
\begin{center}
\includegraphics[width=12cm]{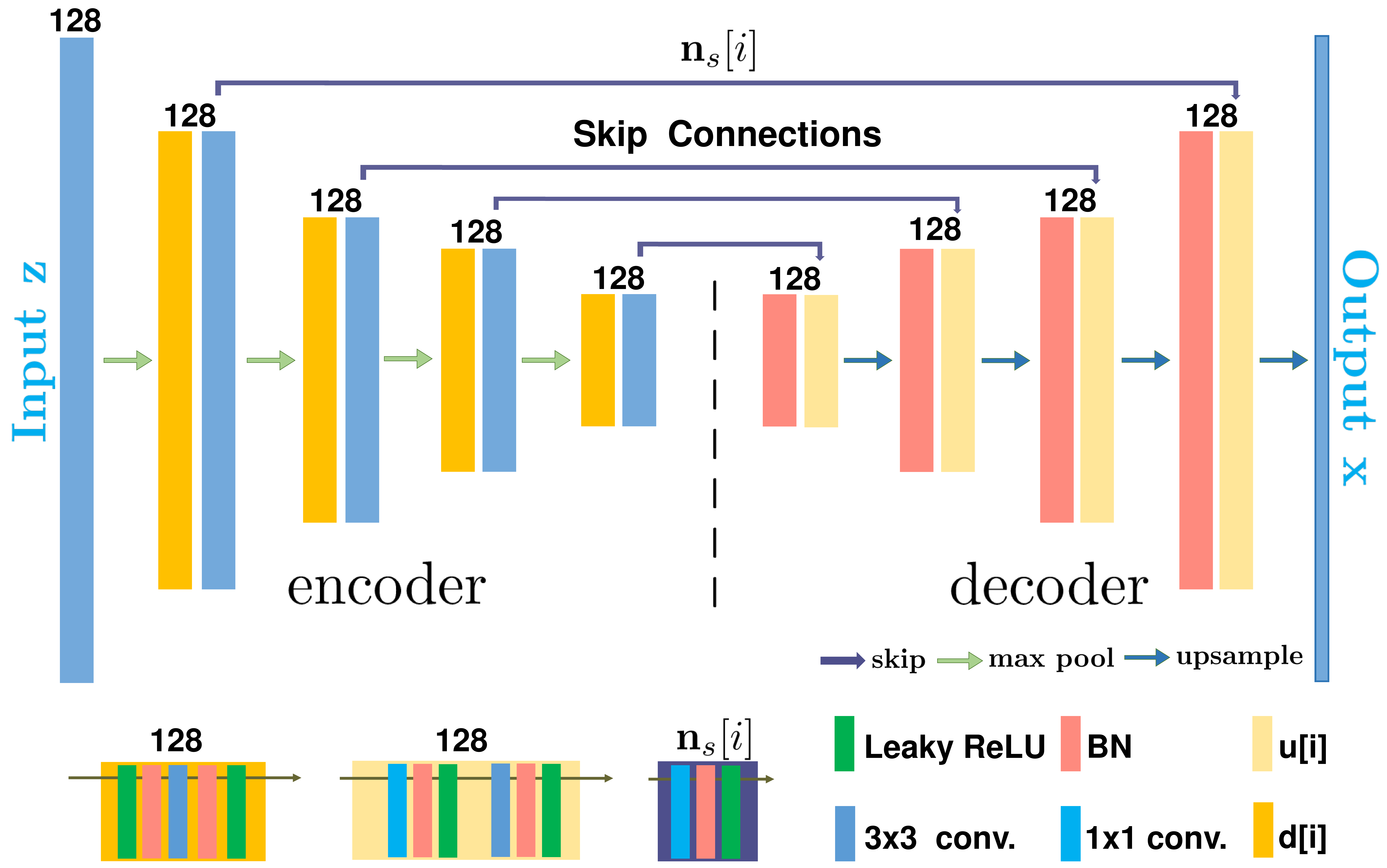}
\end{center}
\caption{CNN architecture~\cite{Ulyanov.etal2018} used in this paper. The architecture is based on the well-known \emph{U-net} with skip connections between the down layers and up layers. Two different kernel sizes are noted under each convolutional layer, and the number of filters is illustrated above each block. The variable $n_s[i]$ denotes the number of feature maps at $i$th skip layer.}
\label{Fig:net_structer}
\end{figure}
The goal of DIP-TV is to use the TV regularization to improve the basic DIP approach. We first consider the optimization problem shown in (\ref{Eq:optimization problem}) and the objective function of DIP in (\ref{Eq:DIP}). One can find that the $\|\ybf-\mathbf{H}f_{\mathbf{\Theta}}(\mathbf{z})\|_{\ell_2}^2$ term in (\ref{Eq:DIP}) actually corresponds to the data-fidelity term in (3) by replacing $f_{\Theta}(\mathbf{z})$ with an unknown image output. Thus, we can consider replacing (\ref{Eq:DIP}) with an optimization problem
\begin{equation}
\label{Eq:DIP-regularization}
\centering
\begin{aligned}
&{\Theta}^* = \argmin_{\Theta}\:\left\{\|\ybf-\mathbf{H}f_{\Theta}(\mathbf{z})\|_{\ell_2}^2+\lambda \rho_{\mathrm{TV}}(f_{\Theta}(\mathbf{z}))\right\},\\
&\text{such that}\quad\xbf^\ast\:=\:f_{\Theta^*}(\mathbf{z}).
\end{aligned}
\end{equation}
Optimization in (\ref{Eq:DIP-regularization}) is similar to training of a CNN and one can rely on any standard optimization algorithms. 

Figure~\ref{Fig:net_structer} illustrates the CNN architecture we used in this paper, which was adapted from~\cite{Ulyanov.etal2018}. In particular, the popular U-net architecture~\cite{ronneberger2015u} is modified such that the skip connections contain a convolutional layer. The decoder uses a down-sampling and up-sampling based scaling-expanding structure, which makes the effective receptive field of the network increase as the input goes deeper into the network~\cite{jin2017deep}. Besides, the skip connection enables the later layers to reconstruct the feature maps with both local details and global texture. Here, the input $\mathbf{z}$ can be initialized with uniform noise and be further optimized. The proposed framework can deal with both grayscale and color images, where for color images anisotropic TV jointly regularizes all three channels.
\begin{figure*}[t]
\begin{center}
\includegraphics[width=16.4cm]{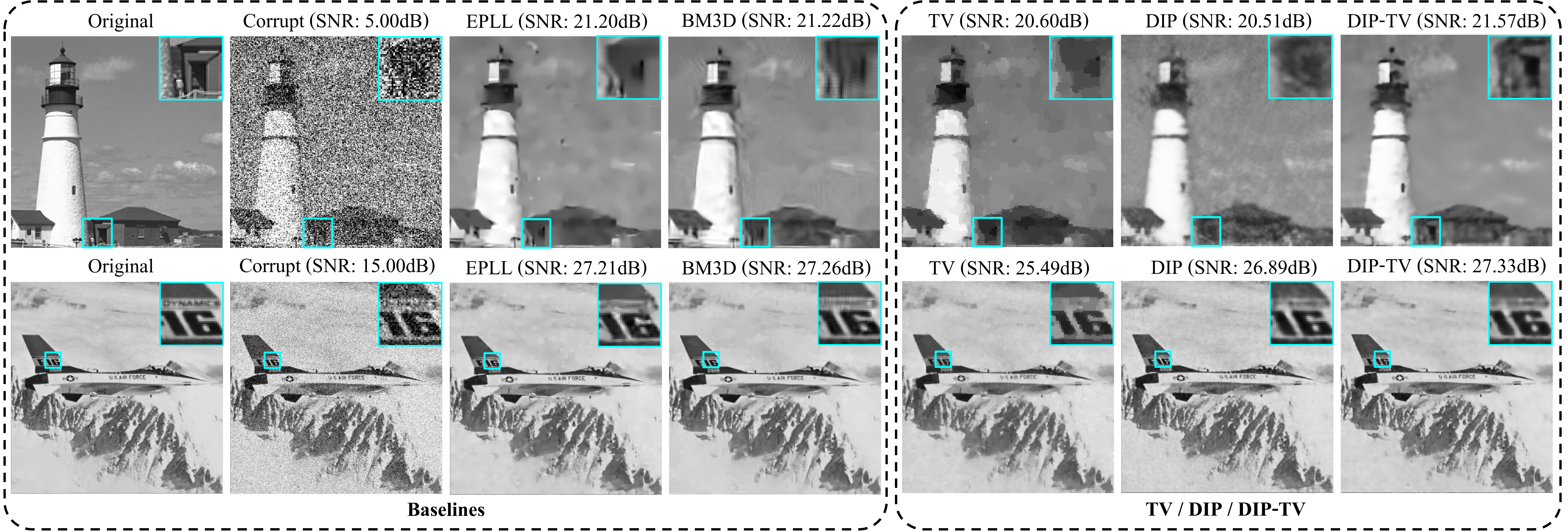}
\end{center}
\caption{Image denoising results on~\emph{Tower} and~\emph{Jet} obtained by EPLL, BM3D, TV-FISTA, DIP, and DIP-TV. The first and second columns display the original images and corrupted images, respectively. Each reconstruction is labeled with its SNR (dB) value with respect to the original image. Visual differences are highlighted using the rectangles drawn inside the images.}
\label{Fig:experiments_gray}
\end{figure*}

\begin{figure*}[t]
\begin{center}
\includegraphics[width=16.3cm]{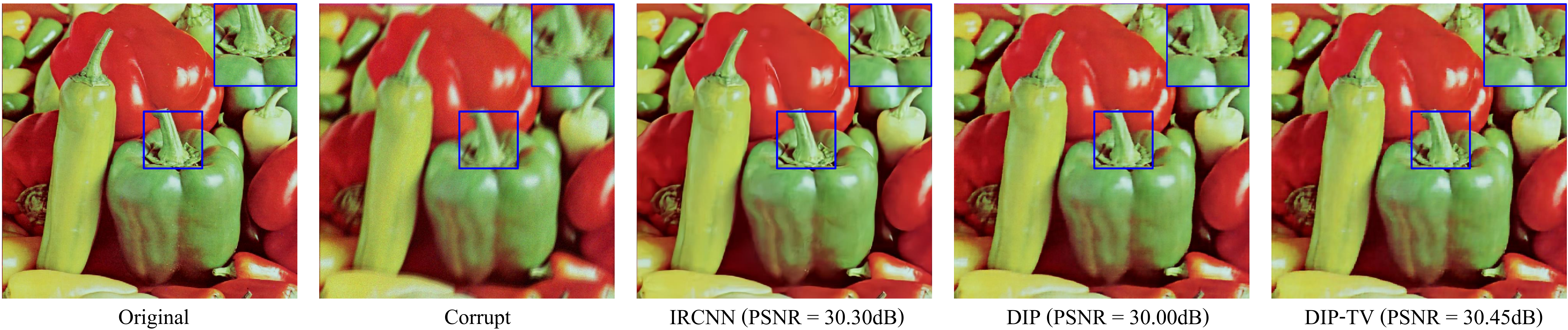}
\end{center}
\caption{Image deblurring results with realistic motion blur kernel from~\cite{levin2009understanding} and $\sigma = 7.65$ on~\emph{Peppers} obtained by IRCNN, DIP, and DIP-TV. Visual differences are highlighted using the rectangles drawn inside the images.}
\label{Fig:experiments_blur}
\end{figure*}
\section{Experiments}
\label{Sec:Experiments}
We now present the experimental results on image denoising and deblurring. We consider 14 gray scale images and 8 standard color images ($256\times256$ and $512\times512$) from set12, set14, and BSD68 as our testing images. The gray scale images are shown in Figure~\ref{Fig:testimages}, while color images are: \emph{Monarch}, \emph{Parrots},  \emph{House}, \emph{Lena}, \emph{Peppers}, \emph{Baby}, and \emph{Jet}.
\begin{table*}[t]
\centering
\textbf{\caption{The SNR (dB) results of different methods on the testing images with input noise levels 5 dB, 10 dB, 15 dB, 20 dB, and 25 dB. For example, 5 dB noisy input represents very high noise level and corresponds to $\sigma=76.26$ in average. \label{Tab:SNR}}}
\scalebox{0.88}{\begin{tabular}{ |c|c|c|c|c|c|c|c|c|c|c|c|c|c|c|}
\hline
\textbf{Images} & 1 & 2& 3 & 4 & 5& 6 & 7 & 8 & 9 & 10 & 11 & 12 & 13 & 14\\
\hline\hline
\multicolumn{15}{|c|}{Input SNR = 5 dB / $\sigma\,=\,$76.26} \\
\hline
EPLL   & 18.60 & 21.39 &   19.18   & 15.29    &16.88&   16.54   & 18.33    & 21.80 &  21.21   & 20.19    &19.38&   19.85 & 16.85 &   21.20\\
\hline
BM3D   & 18.72 & 22.22 &   18.81   & 15.31    &16.86&   16.50   & 18.30    & 21.87 &  21.55   & 20.25    &19.52&   20.35 & 17.33 &   21.22\\
\hline
TV   & 17.22 & 20.38 &   17.65   & 13.74    &16.24&   15.42   & 16.57    & 19.71 &  20.09   & 18.38    & 18.49 &  18.27 & 16.23 &   20.60\\
\hline
DIP    & 17.98 & 21.19 &   18.78   & 14.98    &16.16&   16.19   & 17.61    & 21.44 &  21.08   & 18.67    & 18.97 &   20.19 & 16.64 &   20.51\\
\hline
DIP-TV   & 18.84 & 22.41 &   19.56   & 15.52    &16.99&   16.79   & 18.48    & 22.26 &  21.61   & 19.10    & 19.55 &   20.52 & 17.80 &  21.57\\
\hline
\multicolumn{15}{|c|}{Input SNR = 10 dB / $\sigma\,=\,$53.43} \\
\hline
EPLL   & 21.21 & 24.21 &   21.96   &17.81    &19.42&   19.65   & 20.88    & 24.59 &  23.68   & 21.20    &21.79&   22.98 & 19.65 &   23.91\\
\hline
BM3D   & 21.30 & 25.10 &   21.57 & 17.81 & 19.39 & 19.58 & 20.84 & 24.65 &  24.01   & 21.28 &21.90&   23.39 & 20.20 &   23.85\\
\hline
TV   & 19.76	 & 22.82 &   20.39   & 16.34 &18.45&   18.04   & 18.91    & 22.62 &  22.15   & 20.34    & 20.56 &  20.80 & 18.85 &   22.83\\
\hline
DIP    & 20.76 & 24.32 &   21.55   & 17.81    &18.82&   19.14   & 20.21    & 24.43 &  23.24   & 21.01    & 21.22 &   23.46 & 19.90 &   22.99\\
\hline
DIP-TV   & 21.33 & 25.11 &   22.10   & 17.96    &19.43&   19.61   & 20.89    & 24.77 &  23.81   & 21.57    & 21.65 &   23.60 & 20.46 &  24.12\\
\hline
\multicolumn{15}{|c|}{Input SNR = 15 dB / $\sigma\,=\,$30.02} \\
\hline
EPLL   & 23.57 & 27.04 &   24.63   & 21.00 & 22.10 &   22.79   & 23.12    & 27.21 &  26.29   & 23.65    &24.51&   26.03 & 22.73 &   26.78\\
\hline
BM3D   & 24.02 & 27.95 &   24.55   & 20.96 & 22.04 &   22.69   & 23.41    & 27.26 &  26.60   & 23.71    &24.60&   26.64 & 23.34 &   26.74\\
\hline
TV   & 22.42 & 25.39 &   23.44   & 19.58 & 20.99 &   21.00   & 22.28    & 25.49 &  24.49   & 22.64    & 22.93 &  23.77 & 22.51 & 25.22\\
\hline
DIP    & 23.08 & 26.17 &   23.96  & 20.85 & 21.24 &   22.08   & 22.70    & 26.89 &  25.75   &  22.74    & 23.69 &   26.52 & 22.51 &   25.32\\
\hline
DIP-TV   & 23.77 & 27.37 &   24.63  & 21.05 & 21.85 &   22.59   & 23.12    & 27.33 &  25.97   & 22.90    & 23.95 &   26.81 & 23.22 &  26.65\\
\hline
\multicolumn{15}{|c|}{Input SNR = 20 dB / $\sigma\,=\,$14.24} \\
\hline
EPLL   & 26.59 & 29.26 &   27.35   & 24.19    &24.61&   26.04   &26.41    & 30.11 &  28.78   & 26.50    &27.09&   29.19 & 25.51 &   29.58\\
\hline
BM3D   & 26.78 & 30.20 &   27.36   & 24.16    &24.61&   25.95   & 26.30    & 30.13 &  29.07  & 26.53    &27.14&   29.84 & 26.21 &   29.55\\
\hline
TV   & 25.35 & 27.92 &   26.18   & 23.06    &23.92&   24.34   & 25.13    & 28.42 &  26.99   & 25.36    & 25.60 &  26.94 & 24.97 &   30.86\\
\hline
DIP    & 25.66 & 29.03 &   26.77   & 23.92    &23.94&   25.45   & 25.41    & 29.31 &  27.49   & 23.25    & 25.04 &   29.59 & 25.55 &   28.31\\
\hline
DIP-TV   & 26.37 & 29.53 &   27.38   & 24.10    &24.46&   25.66   & 25.63    & 29.72 &  27.84   & 24.17    & 25.42 &   29.80 & 25.90 &  29.06\\
\hline
\multicolumn{15}{|c|}{Input SNR = 25 dB / $\sigma\,=\,$5.12} \\
\hline
 EPLL   & 30.01 & 31.80 &   30.20   & 27.75    &28.21&   29.51   & 29.51    & 32.86 &  31.11   & 29.58    &29.49&   32.21 & 28.46 &   32.29\\
 \hline
BM3D   & 30.17 & 32.79 &   30.17   & 27.71    &28.17&   29.39   & 29.45    & 32.88 &  31.38   & 29.59    &29.51&   33.00 & 29.12 &   32.27\\
 \hline
 TV   & 28.84 & 30.51 &   29.29   & 26.82    &27.43&   27.90   & 27.81    & 31.36 & 29.77   & 28.45    & 28.47 &  30.42 & 28.24 &   32.63\\
 \hline
DIP    & 28.33 & 31.71 &   29.27   & 26.86 & 26.79 &   28.11   & 27.99    & 30.21 &  27.95   & 24.67   & 25.71 &   31.84 & 28.45 &   30.96\\
\hline
DIP-TV   & 28.75 & 31.80 & 29.92 & 27.42    &26.91&   28.56  & 28.17    & 31.29 &  28.13   & 24.86   & 26.05 &   32.19 & 28.49 &  31.84\\
\hline
\end{tabular}}
\end{table*}
\subsection{Image Denoising}
In this subsection, we analyze the performance of DIP-TV method for image denoising problems. The CNN architecture in Figure~\ref{Fig:net_structer} is used for both color and grayscale images, with $n_s[i] = 4$ for each skip layers. All algorithmic hyperparameters were optimized in each experiment for the best signal-to-noise ratio (SNR) performance with respect to the ground truth test image. Both DIP-TV and DIP were set to run 5000 optimization step. We use the \emph{average SNR} to denote the SNR values averaged over the associated set of test images.

We first present the results of the experiments on grayscale images, where we compared DIP-TV with EPLL~\cite{zoran2011learning}, BM3D~\cite{Dabov.etal2007}, TV~\cite{beck2009fast} and DIP~\cite{Ulyanov.etal2018}. In order to directly evaluate the range of noise levels that DIP-TV performs better, the input SNR to output SNR relationships are presented in Table~\ref{Tab:SNR}. The grayscale images were corrupted by AWGN corresponding to input SNR of 5 dB, 10 dB, 15 dB, 20 dB, 25 dB, respectively. In particular, DIP-TV outperforms original DIP by around 0.5 dB  for a wide range of noise levels from 5 dB to 20 dB. Note that the proposed method also bridge the gap between DIP and the state-of-the-art methods in high noise levels. Figure~\ref{Fig:experiments_gray} illustrates the visual comparisons for grayscale images~\emph{Tower} and~\emph{Jet} under two different noise levels,  respectively. The DIP-TV significantly promotes the denoising performance of DIP itself in terms of both visual qualities and SNR. The noise is effectively filtered out and the details of the image are preserved because of the TV regularization. For instance, DIP-TV improves the SNR with respect to~\emph{Tower} by over 1.06 dB against DIP, and outperforms BM3D by 0.35 dB. Visually, the door highlighted in~\emph{Tower} is clearly restored, while other methods bring serious distortion to it. 

In color image denoising, we compared our method with CBM3D~\cite{Dabov.etal2007} and NLM~\cite{buades2005non} as well as DIP itself. We considered AWGN corresponding to variance $\sigma$ from 25 to 75. Figure~\ref{Fig:color testimage} compares the SNR performance of CBM3D, DIP, and DIP-TV on the image~\emph{Monarch}. Table~\ref{Tab:color SNR} summaries the average SNR among different methods. Overall, DIP-TV exceeds DIP by at least 0.2 dB on the testing images. Moreover, DIP-TV outperforms CBM3D with the increase of noise level (e.g. $\sigma \geq 35$). Considering that the whole procedure of DIP-TV and DIP are image-agnostic and no prior information is learned from other images, it is notable that DIP-TV achieves comparable performance to the state-of-the-art for high noise levels.

\subsection{Image Deblurring}
In image deblurring, one is given an blurry image which is synthesized by firstly applying blur kernel $\mathbf{H}$ and then adding AWGN with noise level $\sigma$; The goal is to restore the image from the degraded ones. We tested DIP and DIP-TV based on the network architecture illustrated in~\cite{Ulyanov.etal2018}, with $n_s[i] = 128$. 
\begin{table*}[t]
\centering
\textbf{\caption{The average SNR (dB) results of CBM3D, NLM, DIP, and DIP-TV on the testing color images with noise level $\sigma$\,=\,25 35 45 55 65 75.\label{Tab:color SNR}}}
\scalebox{1}{
\begin{tabular}{ |c|c|c|c|c|c|c| }
\hline
\textbf{Methods}& $\sigma$\,=\,25 & $\sigma$\,=\,35 & $\sigma$\,=\,45 & $\sigma$\,=\,55 & $\sigma$\,=\,65 & $\sigma$\,=\,75\\
\hline\hline
CBM3D  & \textbf{26.98} & 25.45 & 24.60 & 23.79 & 23.12 & 22.50\\
\hline
NLM & 25.95& 24.19 & 22.97 & 21.83 & 20.90 &20.15\\
\hline
DIP & 26.47	& 25.36 & 24.44 & 23.43 & 22.64 & 22.05\\
\hline
DIP-TV & 26.71 & \textbf{25.50} & \textbf{24.61} & \textbf{23.86} & \textbf{23.21} & \textbf{22.65}\\
\hline
\end{tabular}}
\end{table*}
Both DIP and DIP-TV were set to run 5500 optimization step. Taking advantage of recent progress in CNN and the merit of GPU computation, here we utilized convolution to implement the blur. As a baseline, we compared our method with IRCNN~\cite{Zhang.etal2017a} and DIP itself based on the same set of images in denoising. Two blur kernels were applied, including a general Gaussian kernel with standard deviation 1.6 as well as a realistic kernel defined in ~\cite{levin2009understanding}. Different AWGN of $\sigma$ is added in each experiment.

Figure~\ref{Fig:experiments_blur} shows the visual results for \emph{Peppers} obtained by different methods. All methods can effectively remove the blurry and noise from the image. Particularly, our method further enhance the piecewise-smoothness and mitigate the noise of the image, and thus increases the peak-signal-to-noise ratio (PSNR) by over 0.45 dB against DIP. Also note that the aid of TV regularization makes DIP even outperform IRCNN by 0.15 dB on~\emph{Peppers}. Table~\ref{Tab:PSNR of deblur} reports the average PSNR compassion with IRCNN and DIP on color and gray scale images, repectively. 
\begin{table*}[t]
\centering
\textbf{\caption{The average PSNR (dB) results of IRCNN, DIP and DIP-TV on the testing gray scale images and color images.\label{Tab:PSNR of deblur}}}
\scalebox{1}{
\begin{tabular}{|c|c|c|c|c|}
\hline
\textbf{Methods} & $\sigma$ & IRCNN & DIP & DIP-TV\\
\hline
\multicolumn{5}{|c|}{Gaussian blur with standard deviation 1.6} \\
\hline
\multirow{2}{2em}{Gray Color }&\multirow{2}{*}{2} & 29.76 & 28.65 &  29.44\\ 
& & 32.04 & 31.49 & 32.03\\
\hline
\multicolumn{5}{|c|}{Kernel 1 ($19\times19$~\cite{levin2009understanding})} \\
\hline
\multirow{2}{2em}{Gray Color }&\multirow{2}{*}{2.55}& 32.58 & 31.41 & 32.11\\ 
& & 34.20 & 33.48 & 34.09\\
\hline
\multirow{2}{2em}{Gray Color }&\multirow{2}{*}{7.65} & 28.59 & 26.74 & 27.53\\ 
& & 30.89 & 29.87 & 30.45\\
\hline
\end{tabular}}
\end{table*}
In general, the improvement by TV regularization outperforms DIP by at least 0.54 dB in terms of PSNR and makes the DIP framework more comparable with IRCNN. For example, DIP-TV is only 0.01 dB lower than IRCNN in terms of the average PSNR on color images, with standard Gaussian blur kernel and $\sigma=2$.
\section{Conclusion}
This work has presented a simple method, namely DIP-TV, to improve the deep image prior framework, leading to promising performance, equivalent to and sometimes surpassing recently published leading alternatives, such as BM3D and IRCNN. The proposed method is based on the recent idea that a CNN model itself can act as a prior on images and improve sparsity promoting priors via the $\ell_1$-norm penalty on the image gradient. The results on images denoising and deblurring demonstrate that TV regularization can further improve on DIP and provides high-quality results.


\end{document}